\documentclass{article}

\usepackage [final]{neurips_2021}

\usepackage[utf8]{inputenc} 
\usepackage[T1]{fontenc}    
\usepackage{hyperref}       
\usepackage{url}            
\usepackage{booktabs}       
\usepackage{amsfonts}       
\usepackage{nicefrac}       
\usepackage{microtype}      
\usepackage{xcolor}         
\usepackage{times}  
\usepackage{helvet}  
\usepackage{courier}  
\usepackage{url}  
\usepackage{graphicx}  
\usepackage{bm}
\usepackage{amsfonts,amssymb}
\usepackage{amsmath}
\usepackage{booktabs}
\usepackage{enumerate}
\usepackage{multirow}
\usepackage{subfigure}
\usepackage{epstopdf}
\usepackage{algorithm}
\usepackage{algpseudocode}
\usepackage{amsmath}
\usepackage{graphics}
\usepackage{epsfig}
\usepackage{enumerate}
\usepackage{enumitem}

\newtheorem{theorem}{Theorem}

\newcommand{\plknn}[0]{\textsc{Pl-knn}}
\newcommand{\clpl}[0]{\textsc{Clpl}}
\newcommand{\plsvm}[0]{\textsc{Pl-svm}}

\newcommand{\ipal}[0]{\textsc{Ipal}}
\newcommand{\vpll}[0]{\textsc{Valen}}
\newcommand{\rcpll}[0]{\textsc{RC}}
\newcommand{\ccpll}[0]{\textsc{CC}}
\newcommand{\proden}[0]{\textsc{Proden}}
\newcommand{\dcnn}[0]{\textsc{D2cnn}}
\newcommand{\gaccl}[0]{\textsc{GA}}
\newcommand{\vpllnon}[0]{\textsc{Valen-non}}
\newcommand{\plle}[0]{\textsc{Plle}}

\title{ Instance-Dependent Partial Label Learning}

\author{%
	Ning~Xu, Congyu~Qiao, Xin Geng\thanks{Corresponding author}, and Min-Ling Zhang \\
	School of Computer Science and Engineering, Southeast University, Nanjing 210096, China\\
	MOE Key Laboratory of Computer Network and Information Integration, Ministry of Education, China\\
	\texttt{\{xning, qiaocy, xgeng, zhangml\}@seu.edu.cn} 
}

\begin{document}

\maketitle

\begin{abstract}
Partial label learning (PLL) is a typical weakly supervised learning problem, where each training example is associated with a set of \emph{candidate} labels among which only one is true. Most existing PLL approaches assume that the incorrect labels in each training example are randomly picked as the candidate labels. However, this assumption is not realistic  since the candidate labels are always instance-dependent. In this paper, we consider instance-dependent PLL and assume that each example is associated with a latent \emph{label distribution} constituted by the real number of each label, representing the degree to each label describing the feature. The incorrect label with a high degree  is more likely to be annotated  as the candidate label. Therefore, the latent label distribution is the essential labeling information in partially labeled examples and worth being leveraged for predictive model training. Motivated by  this consideration, we propose a novel PLL method that recovers the label distribution as a  label enhancement (LE) process and trains the predictive model  iteratively in every epoch. Specifically, we assume the true posterior density of the latent label distribution takes on the variational approximate Dirichlet density parameterized by an inference model.  Then  the evidence lower bound is deduced for optimizing the inference model and  the label distributions generated from the variational posterior are utilized for training the predictive model. Experiments on benchmark and real-world datasets validate the effectiveness of the proposed method. Source code is available at \url{https://github.com/palm-ml/valen}.
\end{abstract}

\section{Introduction}\label{sec1}

Partial label  learning (PLL) deals with the problem where each training example is associated with a set of candidate labels, among which only one label is valid \cite{Cour_2011,Chen_2014,Yu_Zhang2017}. Due to the difficulty in collecting exactly labeled data in many real-world
scenarios, PLL leverages    inexact supervision  instead of  exact labels.  The need to learn from the inexact supervision  leads to   a wide range of applications for PLL techniques, such as web mining \cite{Jie_Orabona2010}, multimedia content analysis \cite{Zeng_2013,CPC18}, ecoinformatics \cite{Liu_Dietterich2012,Tang_Zhang2017}, etc.

To accomplish the task of learning from partial label data, many approaches have been proposed. Identification-based PLL approaches \cite{Jin_Ghahramani2003,Nguyen_Caruana2008,Liu_Dietterich2012,Chen_2014,Yu_Zhang2017} regard the ground-truth label  as a latent variable and try to identify it.  Average-based approaches \cite{Hullermeier_Beringer2006,Cour_2011,ZY15} treat all the candidate labels equally and  average the modeling outputs as the prediction. For confidence-based approaches \cite{feng2018leveraging,xu2019partial,zhang2016partial},   the confidence of each label is estimated instead of identifying the  ground-truth label. These approaches  always adopt the randomly picked candidate labels to corrupt benchmark data into partially labeled version  despite having no explicit  generation process of candidate label sets. To  depict the instance-independent  generation process of candidate label sets, Feng \cite{feng2020provably}  proposes a  statistical model  and  deduces  a  risk-consistent method and a  classifier-consistent method. Under the same  generation process, another classifier-consistent risk estimator is proposed for deep model and stochastic optimizers \cite{lv2020progressive}.  

The previous methods assume that the candidate labels are randomly sampled with the uniform generating procedure \cite{lv2020progressive,feng2020provably}, which is commonly adopted to corrupt benchmark datasets into partially labeled versions in their experiments. However, the candidate labels are always instance-dependent (feature-dependent) in practice as the incorrect labels related to the feature  are more likely to  be picked as candidate label set for each instance.  These methods usually do not perform as well as expected  due to the unrealistic assumption on  the generating procedure of candidate label sets.

\begin{figure*}
	\centering
	\subfigure[{ Handwritten digits in MNIST  \cite{lecun1998gradient}}]{\label{fig:MNIST}\includegraphics[width=0.49\textwidth]{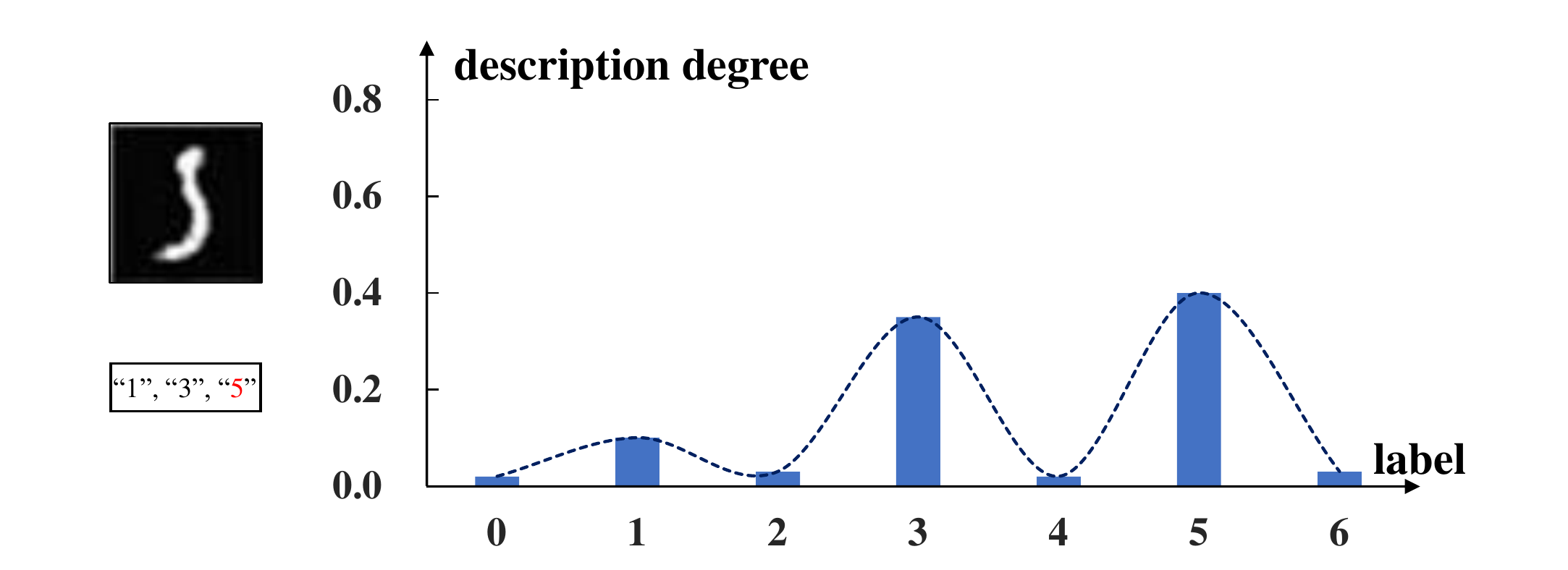}}
	\subfigure[ Color image in CIFAR-10 \cite{krizhevsky2009learning}]{\label{fig:cifar10}\includegraphics[width=0.49\textwidth]{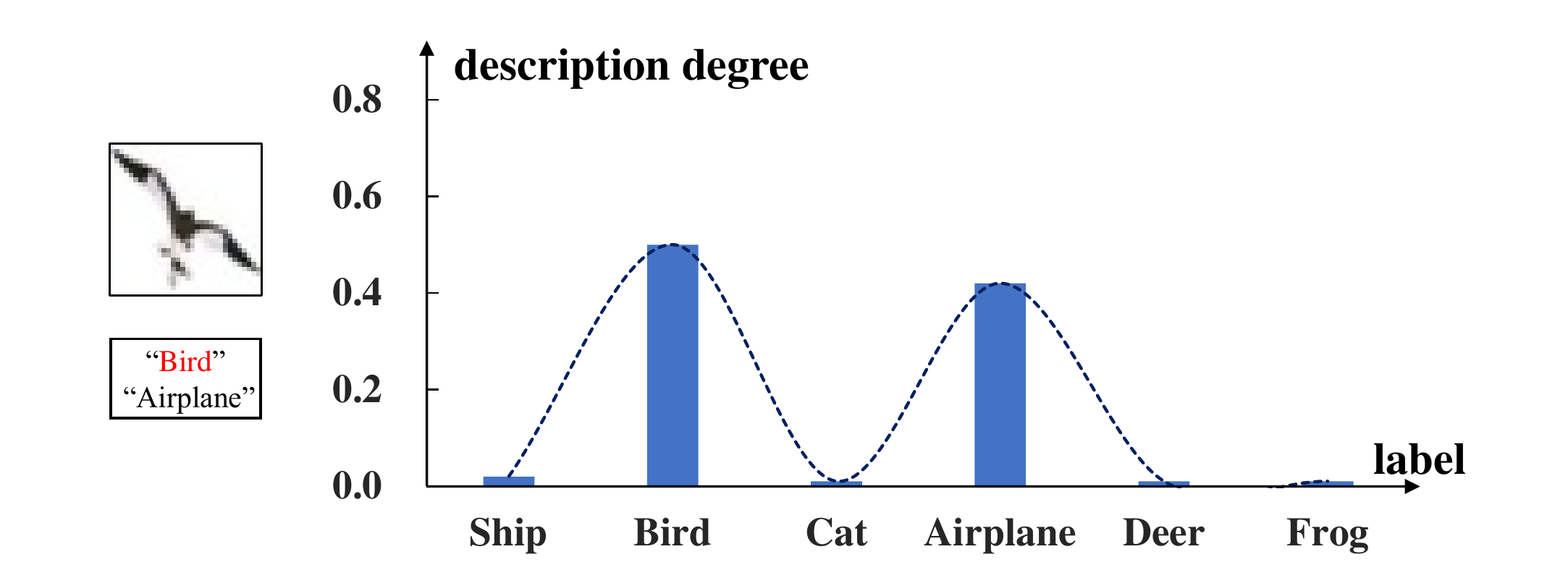}}
	\caption{The examples about the latent label distributions for partial label learning.  The candidate labels are in the box and the red one is valid.}
	\label{fig:relative}
\end{figure*}

In this paper, we consider instance-dependent PLL and assume that each instance in PLL is associated with a latent \emph{label distribution} \cite{xu2019label,xu2020variational,geng2016label} constituted by  the  real number of each label,  representing the degree to each label describing the feature. Then, the incorrect label with a high degree in the latent label distribution  is more likely to be annotated  as the candidate label.  For example, the candidate label set of the handwritten digits in Figure \ref{fig:MNIST}  contains ``1'',  ``3'' and ``{5}'', where  ``1'' and ``3'' are not ground-truth but  selected as candidate labels due to their high degrees in the latent label distribution of the instance. The object in Figure \ref{fig:cifar10} is  annotated with ``bird'' and ``airplane'' as the  degrees of these two labels are much higher than others in the  label distribution. The intrinsical  ambiguity increases the difficulty of  annotating, which leads to the result that  annotators pick the candidate labels with high degrees in the latent label distribution of each instance   instead of annotating the ground-truth label directly in PLL. Therefore, the latent label distribution is the essential labeling information in partially labeled examples and   worth  being leveraged for     predictive model training.

Motivated by the above consideration, we deal with the PLL problem from two aspects. First, we enhance the labeling information by recovering the latent label distribution for each training example as a label enhancement process \cite{xu2019label,xu2020variational}. Second, we run  label enhancement and  train the  predictive model with recovered label distributions  iteratively.  The proposed  method  named {\vpll}, i.e., \emph{VAriational Label ENhancement for instance-dependent partial label learning},  uses the candidate labels to  initialize  the predictive model in the warm-up training stage, then  recovers the latent label distributions  via inferring the variational   posterior  density parameterized by an inference model  with the  deduced evidence lower bound, and trains the predictive model with a risk estimator  by leveraging the candidate labels as well as the label distributions. Our contributions can be summarized as follows:
\begin{itemize}[topsep=0ex,leftmargin=*,parsep=1pt,itemsep=1pt]
	\item  We  for the first time consider the instance-dependent PLL  and assume that each  partially labeled  example is associated with a latent {label distribution}, which is the essential labeling information and worth  being recovered for predictive model training.
	\item  We infer the posterior density of the latent label distribution via taking on the  approximate  Dirichlet density  parameterized by an inference model  and deduce the evidence lower bound  for optimization, in which  the topological information and the features extracted from the predictive model are leveraged.
	\item  We train predictive model with a proposed empirical risk estimator   by leveraging the candidate labels as well as the label distributions. We iteratively  recover the latent label distributions  and train the predictive model  in every epoch.  After the network has been fully trained, the predictive model  can perform	predictions for future test examples alone.
\end{itemize}
 Experiments on the corrupted benchmark datasets and real-world PLL datasets validate the effectiveness of the proposed method.

\section{Proposed Method}

First of all, we briefly introduce some necessary notations. Let $\mathcal{X} = \mathbb{R}^q$ be the $q$-dimensional instance space and $\mathcal{Y}  = \{y_1,y_2,...,y_c\}$ be the label space with $c$ class labels. Given the PLL training set $ \mathcal{D} = \{(\bm{x}_i, S_i) |1 \leq i \leq n\}$ where $\bm{x}_i$ denotes the $q$-dimensional instance and $S_{i} \subseteq \mathcal{Y}$ denotes the candidate label set associated with $\bm{x}_i$. Note that  $S_i$ contains  the correct label of $\bm{x}_i$ and the task of PLL is to induce a multi-class classifier $ f: \mathcal{X} \mapsto \mathcal{Y} $ from $\mathcal{D}$. For each PLL  training example $(\bm{x}_i,S_i)$, we use the logical label vector $\bm{l}_i=[l_i^{y_1},l_i^{y_2},\ldots,l_i^{y_c}]^\top \in \{0,1\}^c$ to represent whether $y_j$ is the candidate label, i.e.,  $l_i^{y_j} = 1$ if $y_j \in S_i$, otherwise $l_i^{y_j} = 0$.  The   label distribution  of $\bm{x}_i$ is denoted by $\bm{d}_i=[d_i^{y_1},d_i^{y_2},\ldots,d_i^{y_c}]^\top \in [0,1]^c$ where $\sum_{j=1}^{c} d_i^{y_j}=1$. Then $\mathbf{L}=[\bm{l}_1,\bm{l}_2,\ldots,\bm{l}_n]$ and $\mathbf{D}=[\bm{d}_1,\bm{d}_2,\ldots,\bm{d}_n]$ represent the logical label matrix and label distribution matrix, respectively.
\subsection{Overview}
To deal with PLL problem,  we iteratively recover the latent label distribution for each example $\bm{x}$ and train the predictive model by leveraging the recovered label distribution.  We start with a warm-up period, in which we train the predictive model with  the PLL  minimal loss \cite{lv2020progressive}. This allows us to attain a
reasonable predictive model before it starts fitting incorrect labels. After the warm-up period,  the features extracted  from the
predictive model can help for recovering the latent label distribution. Benefited from the essential labeling information in the recovered label distribution, the performance of the predictive model could be further improved.

 {\vpll} implements label enhancement and classifier  training  iteratively in every epoch.  In label enhancement, we assume the true posterior density of the latent label distribution takes on the variational approximate Dirichlet density parameterized by an inference model.  Then  the evidence lower bound is deduced for optimizing the inference model and  the label distributions can be generated from the variational posterior.  In  classifier  training, the predictive model is trained by leveraging the recovered label distributions  and candidate labels with an  empirical risk estimator.   After the models has been fully trained, the predictive model  can perform prediction for future test instances alone.
\subsection{Warm-up Training}\label{sec:warm}
The predictive model $\bm{\theta}$ is trained on partially labeled examples  by minimizing the following  PLL minimal loss function \cite{lv2020progressive}:
\begin{equation}\label{minimal}
\mathcal{L}_{min}=\sum_{i=1}^{n}\min_{y_j \in S_i}\ell(f(\boldsymbol{x}_i), \bm{e}^{y_j}),
\end{equation}
where $\ell$ is  cross-entropy loss and $\bm{e}^{\mathcal{Y}} = \{\bm{e}^{y_j}:{y_j} \in \mathcal{Y}\}$ denotes the standard canonical vector
in $\mathcal{R}^c$, i.e., the $j$-element in $\bm{e}^{y_j}$ equals $1$ and others equal $0$.  Similar to \cite{lv2020progressive},  the \texttt{min} operator in Eq. (\ref{minimal}) is replaced by using  the current predictions for slightly  weighting on the  possible labels in   warm-up training. Then we could extract the feature  $\bm{\phi}$ of  each $\bm{x}$  via using the predictive model. 

\subsection{Label Enhancement}\label{sec:recover}
  We assume that the prior density  $ p(\bm{d})$   is a Dirichlet  with  $\hat{\bm{\alpha}}$, i.e., $ p\left(\bm{d}\right)= Dir\left(\bm{d} \mid \hat{\bm{\alpha}}\right)$ where $\hat{\bm{\alpha}} = [\varepsilon,\varepsilon,\ldots,\varepsilon]^\top$ is a $c$-dimensional vector with a minor  value $\varepsilon$. Then we let  the prior density  $ p(\mathbf{D})$  be  the product of each Dirichlet
\begin{equation}\label{prior}
p(\mathbf{D}) = \prod_{i=1}^{n} Dir(\bm{d}_i|\hat{\bm{\alpha}}).
\end{equation}

 We consider the  topological information of the feature space, which is represented by the affinity graph $G=(V,E,{\bf A})$. Here,  the feature vector $\bm{\phi}_i$ of each example could be extracted from the predictive model $\bm{\theta}$ in current epoch, $V=\{{\bm{\phi}}_i \mid 1\leq i\leq n\}$ corresponds to the vertex set consisting of feature vectors,  $E=\{({\bm \phi}_i,{\bm \phi}_j)\mid 1\leq i\neq j\leq n\}$ corresponds to the edge set, and   a sparse adjacency matrix ${\bf A}=[a_{ij}]_{n\times n}$ can be obtained by
\begin{equation}
a_{i j}=\left\{\begin{array}{cl}
1 & \text { if } \bm{\phi}_i \in \mathcal{N}(\bm{\phi}_j) \\
0 &  \text{otherwise}
\end{array}\right.,
\end{equation} 
where $\mathcal{N}(\bm{\phi}_j)$ is the set for  $k$-nearest neighbors of $\bm{\phi}_j$ and the diagonal elements of $\mathbf{A}$ are set to 1.

 Let  features matrix $\mathbf{\Phi}=[\bm{\phi}_1,\bm{\phi}_2,\ldots,\bm{\phi}_n]$, adjacency matrix $\mathbf{A}$ and  logical labels $\mathbf{L}$ be observed matrix, {\vpll} aims to infer the posterior density   $p(\mathbf{D}|\mathbf{L},\mathbf{\Phi},\mathbf{A})$. As the computation of the exact posterior density   $p(\mathbf{D}|\mathbf{L},\mathbf{\Phi},\mathbf{A})$ is intractable,  a fixed-form density  $q(\mathbf{D}|\mathbf{L},\mathbf{\Phi},\mathbf{A})$ is employed to approximate the true posterior. We let  the  approximate posterior be the product of each Dirichlet parameterized by a vector $\bm{\alpha}_i=[\alpha_i^1,\alpha_i^2,\ldots,\alpha_i^c]^\top$:
\begin{equation}\label{posterior}
\begin{split}
q_{\bm{w}}(\mathbf{D} \mid \mathbf{L}, \mathbf{\Phi}, \mathbf{A})=\prod_{i=1}^{n}Dir\left(\bm{d}_i |\boldsymbol{\alpha}_{i}\right).
\end{split}
\end{equation}
Here,  the  parameters  $\mathbf{\Delta}= [\bm{\alpha}_1,\bm{\alpha}_2,\ldots,\bm{\alpha}_n]$ are outputs of the inference model    parameterized by $\bm{w}$, which is defined as a two-layer  GCN \cite{kipf2016variational} by $\mathrm{GCN}(\mathbf{L}, \mathbf{\Phi}, \mathbf{A})=\tilde{\mathbf{A}} \operatorname{ReLU}\left(\tilde{\mathbf{A}} \mathbf{Z} \mathbf{W}_{0}\right) \mathbf{W}_{1}$, with $\mathbf{Z}=[\mathbf{\Phi;L}]$ and weight $\mathbf{W}_{0}$, $\mathbf{W}_{1}$. Here
$\tilde{\mathbf{A}}=\hat{\mathbf{A}}^{-\frac{1}{2}} \mathbf{A} \hat{\mathbf{A}}^{-\frac{1}{2}}$ is the symmetrically normalized weight matrix where $\hat{\mathbf{A}}$ is the degree matrix of ${\mathbf{A}}$.

By following the Variational Bayes techniques, a lower bound on the marginal likelihood of the model is derived which  ensures that $q_{\bm{w}}(\mathbf{D}|\mathbf{L},\mathbf{\Phi},\mathbf{A})$ is as close as possible to $p(\mathbf{D}|\mathbf{L},\mathbf{\Phi},\mathbf{A})$. For logical label  matrix $\mathbf{L}$,   feature matrix $\mathbf{\Phi}$,    and the corresponding $\mathbf{A}$, the log marginal probability can be decomposed as follows \footnote[1]{ More detailed calculations can be seen in  Appendix A.1.}:
\begin{equation}\label{log_marginal}
\begin{split}
\log p(\mathbf{L},\mathbf{\Phi},\mathbf{A}) = \mathcal{L}_{ELBO} +\text{KL}[q_{\bm{w}}(\mathbf{D}|\mathbf{L},\mathbf{\Phi},\mathbf{A})||p(\mathbf{D}|\mathbf{L},\mathbf{\Phi},\mathbf{A})].
\end{split}
\end{equation}
where
\begin{equation}\label{lowerbound}
\begin{split}
\mathcal{L}_{ELBO} = \mathbb{E}_{q_{\bm{w}}({\bf D}|{\bf L}, {\mathbf{\Phi}}, {\bf A})}[ \log p({\bf L}, {\mathbf{\Phi}}, {\bf A}|{\bf D})]- \text{KL}[ q_{\bm{w}}({\bf D}|{\bf L}, {\mathbf{\Phi}}, {\bf A})|| p({\bf D})].
\end{split}
\end{equation} 
Due to the non-negative property of KL divergence, the
first term $\mathcal{L}_{ELBO}$ constitutes a lower bound of
$\log p(\mathbf{L},\mathbf{\Phi},\mathbf{A})$, which is often called as evidence lower bound
(ELBO), i.e., $\log p(\mathbf{L},\mathbf{\Phi},\mathbf{A}) \geq \mathcal{L}_{ELBO}$.

 According to Eq. (\ref{prior}) and Eq. (\ref{posterior}) , the   KL divergence in  Eq. (\ref{lowerbound}) can be  analytically calculated as follows:
\begin{equation}\label{D_KL}
\begin{array}{l}
\operatorname{KL}\left(q_{\bm{w}}(\mathbf{D}|\mathbf{L},\mathbf{\Phi},\mathbf{A}) \| p(\mathbf{D})\right)= \sum_{i=1}^{n}\bigg (\log \Gamma\left(\sum_{j=1}^{c} \alpha_{i}^j\right) 
-\sum_{j=1}^{c} \log \Gamma\left(\alpha_{i}^j\right)\\-\log \Gamma\left(c\cdot\varepsilon\right)+c\log \Gamma\left
(\varepsilon\right) 
+\sum_{j=1}^{c}\left(\alpha_{i}^j-\varepsilon\right)\left(\psi\left(\alpha_{i}^j\right)-\psi\left(\sum_{j=1}^{c} \alpha_{i}^j\right)\right) \bigg ).
\end{array}
\end{equation}
where $\Gamma(\cdot)$ and $\psi(\cdot)$ are  Gamma function and Digamma function, respectively.

As the first part of Eq. (\ref{lowerbound})  is intractable, we employ the  implicit reparameterization trick \cite{figurnov2018implicit} to approximate it by Monte Carlo (MC) estimation. Inspired by \cite{kipf2016variational},  we simply drop the dependence
on $\mathbf{\Phi}$:
\begin{equation}
\begin{split}
&p(\mathbf{L} \mid {\bf A}, \mathbf{D})=\prod_{i=1}^{n} p\left(\boldsymbol{l}_{i} \mid {\bf A},\mathbf{D}\right), \\
&p(\mathbf{A} \mid \mathbf{D})=\prod_{i=1}^{{n}} \prod_{j=1}^{n} p\left(a_{i j} \mid \bm{d}_{i}, \bm{d}_{j}\right), \text { with } p\left(a_{i j}=1 \mid \bm{d}_{i}, \bm{d}_{j}\right)=s\left(\bm{d}_{i}^{\top} \bm{d}_{j}\right).
\end{split}\end{equation}
Here, $s(\cdot)$ is the logistic sigmoid function. We further assume  that $p\left(\bm{l}_{i}| {\bf A},\mathbf{D}\right)$ is a multivariate Bernoulli with probabilities $\bm{\tau}_i$.  In order to simplify the observation model,  $\mathbf{T}^{(m)}=[\bm{\tau}_1^{(m)}, \bm{\tau}_2^{(m)},\ldots,\bm{\tau}_n^{(m)}]$ is computed from $m$-th sampling $\mathbf{D}^{(m)}$ with a three-layer MLP parameterized by $\bm{\eta}$.  Then the first part of Eq. (\ref{lowerbound}) can be tractable:

\begin{equation}\label{part1}
\begin{split}
\mathbb{E}_{q_{\bm{w}}({\bf D}|{\bf L}, {\mathbf{\Phi}}, {\bf A})}[ \log p_{\bm{\eta}}({\bf L}, {\mathbf{\Phi}}, {\bf A}|{\bf D})] = 
 \frac{1}{M}\sum_{m=1}^{M} \Bigg  (  \operatorname{tr}\left( \left(\mathbf{I}-\mathbf{L}\right)^\top \log \left(\mathbf{I}-\mathbf{T}^{(m)}\right) \right)    
 \\+\operatorname{tr}\left( \mathbf{L}^{\top} \log \mathbf{T}^{(m)} \right)- \Vert \mathbf{A}-S\left(\mathbf{D}^{(m)}\mathbf{D}^{(m)\top}\right) \Vert_F^2 \Bigg ) . 
\end{split}
\end{equation}
Note that we can use only one MC sample in Eq. (\ref{part1}) during the training process as suggested in \cite{kingma2014auto,xu2020variational}.

In addition, {\vpll}  improves the label enhancement  by employing the  compatibility loss, which  enforces that the recovered label distributions should not be completely different from the  confidence $\zeta(\bm{x}_i)$ \cite{feng2020provably,lv2020progressive} estimated by current prediction $f(\bm{x}_i;\bm{\theta})$:
\begin{equation}
\mathcal{L}_{o}=-\frac{1}{n} \sum_{i=1}^{n} \sum_{j=1}^{c} \zeta_j(\bm{x}_i)\log d_{i}^{y_j}
\end{equation}
where \begin{equation}
	\zeta_j(\bm{x}_i)=\left\{\begin{array}{cl}
		f_{j}\left(\boldsymbol{x}_{i};\bm{\theta}\right) / \sum_{y_k \in S_{i}} f_{k}\left(\boldsymbol{x}_{i};\bm{\theta}\right) & \text { if } y_j \in S_{i} \\
		0 & \text { otherwise }
	\end{array}\right.
\end{equation}
 Now we can easily get the  objective  of label enhancement $\mathcal{L}_{LE}$ as follows:
\begin{equation}\label{target_enhancement}
\mathcal{L}_{LE}  = \lambda\mathcal{L}_{o}- \mathcal{L}_{ELBO}
\end{equation}
where $\lambda$ is a hyper-parameter. The label distribution matrix $\mathbf{D}$ is sampled from $q(\mathbf{D}|\mathbf{L},\mathbf{\Phi},\mathbf{A})$, i.e., $\bm{d}_i \sim Dir(\bm{\alpha}_i)$.  Note that the implicit reparameterization gradient \cite{figurnov2018implicit} is employed, which avoids the inversion of the standardization function, which makes the gradients can be computed  analytically in backward pass. 
\begin{algorithm}[t] 
	\caption{ {\vpll} Algorithm} 
	\label{alg:Framwork} 
	\begin{algorithmic}[1] 
		\Require 
		The PLL training set $\mathcal{D} = \{(\bm{x}_i,S_i)\}_{i=1}^n$, epoch $T$ and iteration $I$;
		\State Initialize the predictive model $\bm{\theta}$ by warm-up training,  the reference model $\bm{w}$ and observation model $\bm{\eta}$;
			\State Extract the features $\mathbf{\Phi}$ from predictive model $\bm{\theta}$ and calculate the adjacency matrix $\mathbf{A}$;
	
		\For {$t=1,\ldots, T$}

		\State Shuffle training set $\mathcal{D} = \{(\bm{x}_i,S_i)\}_{i=1}^n$ into $I$ mini-batches;
		\For {$k=1,\ldots, I$}
		\State Obtain label distribution   $\bm{d}_i$ for each example $\bm{x}_i$ by Eq. (\ref{posterior}); 
		\State Update  $\bm{\theta}$, $\bm{w}$ and $\bm{\eta}$ by forward computation and back-propagation by fusing Eq. (\ref{target_enhancement}) and Eq. (\ref{estimator}); 
		\EndFor
		\EndFor
		\Ensure The predictive model $\bm{\theta}$.
	\end{algorithmic} 
\end{algorithm}

\subsection{Classifier Training}
 To train the predictive model, we minimize the following empirical risk estimator by levering   the recovered label distributions:
 \begin{equation}\label{estimator}
	\widehat{R}_V(f) = \frac{1}{n} \sum_{i=1}^{n}\left(\sum_{y_j \in S_i}\frac{d_i^{y_j}}{\sum_{y_j\in S_i}d_i^{y_j}}\ell(f(\bm{x}_i),\bm{e}^{y_j})\right).
\end{equation}
Here we adopt the average value of  $\bm{d}_i$  sampled by $\bm{d}_i \sim Dir(\bm{\alpha}_i)$. We can use any deep neural network as the predictive model, and then equip it with the {\vpll} framework to deal with PLL.  Note that we could train the predictive model  and update the label distributions in a principled end-to-end manner by fusing the objective Eq. (\ref{target_enhancement}) and Eq. (\ref{estimator}).  The algorithmic description of the {\vpll}  is shown in Algorithm 1.

Let $\widehat{f}_{V}=\min _{f \in \mathcal{F}} \widehat{R}_{V}(f)$ be the empirical risk minimizer and $f^{\star}=\min _{f \in \mathcal{F}} R_V(f)$
be the optimal risk minimizer where $ R_V(f)$ is the risk estimator. Besides, we define the function space $\mathcal{H}_{y_j}$ for the label $y_j \in \mathcal{Y}$ as $\left\{h: \boldsymbol{x} \mapsto f_{y_j}(\boldsymbol{x}) \mid f \in \mathcal{F}\right\}$. Let $\mathfrak{R}_{n}\left(\mathcal{H}_{y_j}\right)$ be the expected Rademacher complexity \cite{bartlett2002rademacher} of $\mathcal{H}_{y_j}$ with sample size $n$, then we have the following theorem.
\begin{theorem}\label{theorem}
	Assume the loss function $\ell(f(\boldsymbol{x}), \bm{e}^{y_j})$ is $L$-Lipschitz with respect to $f(\boldsymbol{x})(0<L<\infty)$ for all $y_j \in \mathcal{Y}$ and upper-bounded by $M$, i.e., $M=\sup _{x \in \mathcal{X}, f \in \mathcal{F}, y_j \in \mathcal{Y}} \ell(f(x), \bm{e}^{y_j})$. Then, for any
	$\delta>0$, with probability at least $1-\delta$,
	$$
	R\left(\widehat{f}_{V}\right)-R\left(f^{\star}\right) \leq 4 \sqrt{2} L \sum_{j=1}^{c} \mathfrak{R}_{n}\left(\mathcal{H}_{y_j}\right)+M \sqrt{\frac{\log \frac{2}{\delta}}{2 n}}
	$$
\end{theorem}
The proof of Theorem \ref{theorem} is provided in Appendix A.2.   Theorem \ref{theorem} shows that the empirical risk minimizer $f_{V}$ converges to the optimal risk minimizer $f^{\star}$ as $n \rightarrow \infty$ and $\mathfrak{R}_{n}\left(\mathcal{H}_{y_j}\right) \rightarrow 0$ for all parametric models with a bounded norm.

\section{Related Work}
As shown in Section \ref{sec1}, supervision information conveyed by partially labeled training examples is implicit as the ground-truth label is hidden within the candidate label set. Therefore, partial label learning can be regarded as a \emph{weak supervision} learning framework \cite{katz2019decontamination} with implicit labeling information. Intuitively, the basic strategy for handling partial label learning  is disambiguation, i.e., trying to identify the ground-truth label from the candidate label set associated with each training example, where existing strategies include disambiguation by identification or disambiguation by averaging. For identification-based disambiguation, the ground-truth label is regarded as a latent variable and identified \cite{Jin_Ghahramani2003,Nguyen_Caruana2008,Liu_Dietterich2012,Chen_2014,Yu_Zhang2017}. For averaging-based disambiguation, all the candidate labels are treated equally and the prediction is made by averaging their modeling outputs \cite{Hullermeier_Beringer2006,Cour_2011,ZY15}.

Most existing algorithms aim to fulfill the learning task by fitting widely-used learning techniques to partial label data. For maximum likelihood techniques, the likelihood of observing each partially labeled training example is defined over its candidate label set instead of the unknown ground-truth label \cite{Jin_Ghahramani2003,Liu_Dietterich2012}. $K$-nearest neighbor techniques determine  the class label of unseen instances via voting among the candidate labels of its neighboring examples \cite{Hullermeier_Beringer2006,ZY15}. For maximum margin techniques, the classification margins over the partially labeled training examples are defined by discriminating modeling outputs from candidate labels and non-candidate labels \cite{Nguyen_Caruana2008,Yu_Zhang2017}. For boosting techniques, the weight over each partially labeled training example and the confidence over the candidate labels are updated in each boosting round \cite{Tang_Zhang2017}. For disambiguation-free strategies,  the generalized  description degree is estimated by using a graph Laplacian and induce a multi-output regression \cite{xu2019partial}. The confidence of each candidate label is estimated by using the  manifold structure of feature space \cite{zhang2016partial}. However,   these methods just estimate the soft labeling information and train  the predictive models in separate stages without considering the  feedback of the predictive models.

The above-mentioned works were solved in specific low-efficiency manners and incompatible with high-efficient stochastic optimization. To   handle large-scale datasets,   the deep networks are employed with an entropy-based regularizer to maximize the margin between the potentially correct label and the unlikely ones \cite{yao2020deep}. \cite{lv2020progressive} proposes a classifier-consistent risk estimator and a progressive identification, which is compatible with deep models and stochastic optimizers.  \cite{feng2020provably} proposes  a  statistical model to depict the generation process of candidate label sets, which deduces  a  risk-consistent method and a  classifier-consistent method.

The previous methods assume that the candidate labels are randomly sampled with the uniform generating procedure. However, the candidate labels are always instance-dependent (feature-dependent) in practice as the incorrect labels related to the feature  are more likely to  be picked as candidate label set for each instance.  
In this paper, we consider instance-dependent PLL and assume that each instance in PLL is associated with a latent \emph{label distribution} \cite{xu2019label,xu2020variational,geng2016label} constituted by  the  real number of each label,  representing the degree to each label describing the feature.  Label enhancement (LE)  \cite{geng2013facial,xu2019label,xu2020variational}  recovers the latent label distribution from the observed logical labels, in which the recovered label distribution is a kind of pseudo label \cite{pham2021meta,lee2013pseudo}, actually. Note that the generation process of  soft labels in label smoothing \cite{szegedy2016rethinking,muller2019does} and distillation \cite{hinton2015distilling,zhang2019your}   could also be regarded as a label enhancement process.

\section{Experiments}

\subsection{Datasets}

We adopt four widely used benchmark datasets including \texttt{MNIST} \cite{lecun1998gradient}, \texttt{Fashion-MNIST} \cite{xiao2017fashion}, \texttt{Kuzushiji-MNIST} \cite{clanuwat2018deep}, and \texttt{CIFAR-10} \cite{krizhevsky2009learning},
and five datasets from the UCI Machine Learning Repository \cite{Bache_Lichman2013}, including \texttt{Yeast}, \texttt{Texture}, \texttt{Dermatology}, \texttt{Synthetic Control}, and \texttt{20Newgroups}. 

We manually corrupt these datasets into partially labeled versions \footnote{The datasets corrupted by the instance-dependent generating procedure are available at \url{https://drive.google.com/drive/folders/1J_68EqOrLN6tA56RcyTgcr1komJB31Y1?usp=sharing}.} by using a flipping probability $\xi_i^{y_j}=P({l}_i^{y_j}=1|\hat{l}_i^{y_j}=0,\bm{x}_i)$ , where $\hat{l}_i^{y_j}$ is the original clean label. To synthesize the instance-dependent candidate labels, we  set the flipping probability of each incorrect label corresponding to an example $\bm{x}_i$ by using the confidence prediction of a clean neural network $\hat{\bm{\theta}}$ (trained with the original clean labels) \cite{zhang2021learning} with $\xi_i^{y_j}=\frac{f_j(\bm{x}_i;\hat{\bm{\theta}})}{\max_{{y}_j \in \bar{Y}_i}f_j(\bm{x}_i;\hat{\bm{\theta}})}$, where $\bar{Y}_i$ is the incorrect label set of $\bm{x}_i$.  The uniform corrupted version adopts the uniform generating procedure \cite{lv2020progressive,feng2020provably}  to flip the incorrect label into candidate label, where $\xi_i^{y_j} = \frac{1}{2}$.  

\begin{table}
	\caption{Classification accuracy (mean$\pm$std) of each comparing approach on  benchmark datasets corrupted by the instance-dependent generating procedure. }
	\centering
	\label{bench_feature}
	\begin{tabular}{ccccc}
		\toprule
		& MNIST & Kuzushiji-MNIST & Fashion-MNIST & CIFAR-10 \\
		\midrule
		{\vpll}    & \textbf{97.85$\pm$0.05\%} & \textbf{86.19$\pm$0.14\%} & \textbf{86.17$\pm$0.19\%} & \textbf{80.38$\pm$0.52\%} \\
		\midrule
		{\proden}  & 97.69$\pm$0.04\%$\bullet$ & 85.71$\pm$0.12\%$\bullet$ & 85.54$\pm$0.09\%$\bullet$ & 79.80$\pm$0.28\%$\bullet$ \\		
		{\rcpll}   & 97.60$\pm$0.05\%$\bullet$ & 84.86$\pm$0.11\%$\bullet$ & 85.51$\pm$0.10\%$\bullet$ & 79.46$\pm$0.25\%$\bullet$ \\
		{\ccpll}   & 97.44$\pm$0.03\%$\bullet$ & 82.67$\pm$1.82\%$\bullet$ & 85.19$\pm$0.04\%$\bullet$ & 78.98$\pm$0.60\%$\bullet$ \\
		{\dcnn}    & 94.63$\pm$0.16\%$\bullet$ & 83.03$\pm$0.78\%$\bullet$ & 82.42$\pm$0.21\%$\bullet$ & 73.11$\pm$0.11\%$\bullet$ \\
		{\gaccl}   & 95.25$\pm$0.07\%$\bullet$ & 82.45$\pm$0.63\%$\bullet$ & 80.41$\pm$0.24\%$\bullet$ & 77.57$\pm$0.76\%$\bullet$ \\
		\bottomrule
	\end{tabular}
\end{table}

\begin{table}[t]
	\caption{ Classification accuracy (mean$\pm$std) of each comparing approach on  benchmark datasets corrupted by  the uniform generating procedure. }
	\label{benchmark_uniform}
	\centering
	\begin{tabular}{cccccc}
		\toprule
		& MNIST & Kuzushiji-MNIST & Fashion-MNIST & CIFAR-10 \\
		\midrule
		{\vpll}    & {97.93$\pm$0.05\%} & \textbf{88.76$\pm$0.26\%} & \textbf{88.98$\pm$0.16\%} & \textbf{81.93$\pm$1.01\%} \\
		\midrule
		{\proden}  & \textbf{97.97$\pm$0.03\%} & 88.55$\pm$0.10\% & 88.94$\pm$0.12\% & 81.53$\pm$0.53\% \\
		{\rcpll}   & 97.86$\pm$0.03\% & 86.65$\pm$0.10\%$\bullet$ & 88.59$\pm$0.08\%$\bullet$ & 81.30$\pm$1.30\% \\
		{\ccpll}   & 97.73$\pm$0.02\%$\bullet$ & 87.99$\pm$0.03\%$\bullet$ & 88.93$\pm$0.06\% & 80.17$\pm$1.09\%$\bullet$ \\
		{\dcnn}    & 95.12$\pm$0.16\%$\bullet$ & 84.03$\pm$0.78\%$\bullet$ & 80.42$\pm$0.21\%$\bullet$ & 75.11$\pm$0.11\%$\bullet$ \\
		{\gaccl}   & 96.29$\pm$0.19\%$\bullet$ & 82.36$\pm$0.98\%$\bullet$ & 81.81$\pm$0.99\%$\bullet$ & 60.14$\pm$1.35\%$\bullet$ \\

		\bottomrule
	\end{tabular}
\end{table}

\begin{table}[t]
	\caption{Classification accuracy (mean$\pm$std) of each comparing approach on  UCI datasets corrupted by  the instance-dependent generating procedure. }
	\centering
	\small
	\label{UCI_feature}
	\begin{tabular}{cccccc}
		\toprule
		&  Yeast & Texture & Synthetic Control & Dermatology & 20Newsgroup \\
		\midrule
		{\vpll}     & \textbf{57.57$\pm$1.08\%} & \textbf{94.76$\pm$0.93\%} & \textbf{82.86$\pm$0.76\%} & 89.86$\pm$1.31\% & \textbf{81.88$\pm$0.47\%} \\
		\midrule
		{\proden}   & 54.78$\pm$1.28\%$\bullet$ & 89.87$\pm$2.14\%$\bullet$ & 71.16$\pm$6.19\%$\bullet$ & 88.53$\pm$3.87\% & 78.06$\pm$0.74\%$\bullet$\\		
		{\rcpll}    & 54.77$\pm$1.27\%$\bullet$ & 89.57$\pm$2.37\%$\bullet$ & 65.99$\pm$2.72\%$\bullet$ & 88.53$\pm$3.49\% & 78.02$\pm$0.79\%$\bullet$\\
		{\ccpll}    & 54.98$\pm$0.91\%$\bullet$ & 88.92$\pm$7.56\%
		& 66.99$\pm$2.47\%$\bullet$ & 88.26$\pm$4.11\%  & 77.88$\pm$0.39\%$\bullet$\\
		{\dcnn}     & 44.94$\pm$1.87\%$\bullet$ & 69.52$\pm$5.79\%$\bullet$ & 62.66$\pm$8.92\%$\bullet$ & 81.95$\pm$6.18\%$\bullet$  & 73.55$\pm$0.92\%$\bullet$\\
		{\gaccl}    & 25.86$\pm$3.17\%$\bullet$ & 74.84$\pm$2.87\%$\bullet$ & 56.43$\pm$1.29\%$\bullet$ & 84.85$\pm$1.43\%$\bullet$  & 49.49$\pm$3.42\%$\bullet$\\
		\midrule
		{\clpl}     & 54.92$\pm$2.38\% & 81.27$\pm$9.09\%$\bullet$ & 66.33$\pm$3.25\%$\bullet$ & 92.07$\pm$3.42\% & 77.62$\pm$0.23\%$\bullet$ \\
		{\plsvm}    & 41.85$\pm$5.92\%$\bullet$ & 39.03$\pm$4.35\%$\bullet$ & 50.33$\pm$5.73\%$\bullet$ & 84.98$\pm$4.56\% & 72.89$\pm$0.41\%$\bullet$ \\
		{\plknn}    & 47.44$\pm$2.69\%$\bullet$ & 70.05$\pm$0.70\%$\bullet$ & 80.50$\pm$1.26\%$\bullet$ & 83.61$\pm$3.15\%$\bullet$ & 33.28$\pm$1.09\%$\bullet$ \\
		{\ipal}     & 56.40$\pm$2.07\% & 93.49$\pm$0.89\% & 
		77.66$\pm$3.60\%$\bullet$ & 78.94$\pm$8.34\%$\bullet$ & 67.38$\pm$0.95\%$\bullet$ \\
		{\plle}     & 55.53$\pm$1.74\% & 84.45$\pm$1.07\%$\bullet$ & 
		66.16$\pm$7.96\%$\bullet$ & \textbf{93.16$\pm$2.58\%}$\circ$ & 75.54$\pm$0.66\%$\bullet$ \\
		\bottomrule
	\end{tabular}
\end{table}

In addition, five real-world PLL datasets are adopted, which are collected from several application domains including \texttt{Lost} \cite{Cour_2011}, \texttt{Soccer Player} \cite{Zeng_2013} and \texttt{Yahoo!News} \cite{Guillaumin_2010} for automatic face naming from images or videos, \texttt{MSRCv2} \cite{Liu_Dietterich2012} for object classification, and \texttt{BirdSong} \cite{Briggs_2013} for bird song classification.  The detailed descriptions of these
datasets   are provided in Appendix A.3.

We run 5 trials on the four benchmark datasets and perform five-fold cross-validation on UCI datasets and real-world PLL datasets. The mean accuracy as well as standard deviation are recorded for all comparing approaches.

\begin{table}[t]
	\caption{Classification accuracy (mean$\pm$std) of each comparing approach on  UCI datasets corrupted by  the uniform generating procedure. }
	\label{UCI_uniform}
	\centering
	\small
	\begin{tabular}{cccccc}
		\toprule
		&  Yeast & Texture & Synthetic Control & Dermatology  & 20Newsgroup \\
		\midrule
		{\vpll}     & \textbf{58.18$\pm$1.46\%} & 97.30$\pm$0.57\% & \textbf{97.17$\pm$0.47\%} & \textbf{97.07$\pm$0.41\%} & \textbf{71.75$\pm$3.02\%} \\
		\midrule
		{\proden}   & 56.32$\pm$1.98\% & 97.75$\pm$0.53\% & 95.83$\pm$1.95\% & 95.07$\pm$1.84\%$\bullet$ & 68.28$\pm$0.91\%$\bullet$ \\
		{\rcpll}    & 56.39$\pm$1.85\% & 97.77$\pm$0.55\% & 95.99$\pm$1.80\% & 95.62$\pm$1.51\% & 68.44$\pm$1.09\%  \\
		{\ccpll}    & 56.25$\pm$1.89\% & 97.79$\pm$0.57\% & 96.33$\pm$1.39\% & 95.90$\pm$1.69\% & 67.95$\pm$0.95\%$\bullet$ \\
		{\dcnn}     & 54.04$\pm$1.90\%$\bullet$ & 97.23$\pm$0.72\% & 81.16$\pm$8.11\%$\bullet$ & 90.43$\pm$2.38\%$\bullet$  & 65.88$\pm$2.56\%$\bullet$\\
		{\gaccl}    & 22.98$\pm$2.57\%$\bullet$ & 95.09$\pm$1.07\%$\bullet$ & 56.87$\pm$1.53\%$\bullet$ & 51.95$\pm$3.89\%$\bullet$  & 58.29$\pm$1.74\%$\bullet$\\
		\midrule
		{\clpl}     & 56.54$\pm$3.35\% & 98.14$\pm$0.59\% & 94.66$\pm$6.41\% & 96.72$\pm$0.76\% & 70.45$\pm$0.91\% \\
		{\plsvm}    & 46.23$\pm$7.21\%$\bullet$ & 39.74$\pm$2.11\%$\bullet$ & 76.50$\pm$5.31\%$\bullet$ & 92.37$\pm$5.08\% & 70.44$\pm$0.37\% \\
		{\plknn}    & 44.40$\pm$2.50\%$\bullet$ & 95.31$\pm$0.85\%$\bullet$ & 95.33$\pm$2.98\% & 92.91$\pm$2.92\%$\bullet$ & 27.10$\pm$0.49\%$\bullet$ \\
		{\ipal}     & 43.86$\pm$3.39\%$\bullet$ & \textbf{98.71$\pm$0.37\%}$\circ$ & 96.83$\pm$1.90\% & 95.35$\pm$2.08\% & 65.39$\pm$1.21\%$\bullet$ \\
		{\plle}     & 53.58$\pm$2.86\%$\bullet$ & 98.40$\pm$0.40\%$\circ$ & 89.66$\pm$1.91\%$\bullet$ & 90.98$\pm$1.85\%$\bullet$ & 53.88$\pm$0.59\%$\bullet$ \\
		\bottomrule
	\end{tabular}
\end{table}

\begin{table}[t]
	\caption{Classification accuracy (mean$\pm$std) of each comparing approach on  the real-world datasets.  }
	\label{real-world datasets}
	\centering
	\small
	\begin{tabular}{cccccc}
		\toprule
		&  Lost    & MSRCv2    & BirdSong            & Soccer Player     &     Yahoo!News \\
		\midrule
		{\vpll}     & 70.28$\pm$2.29\%  & 47.61$\pm$1.79\% & \textbf{72.02$\pm$0.37\%}& \textbf{55.90$\pm$0.58\%} & \textbf{67.52$\pm$0.19\%} \\
		\midrule
		{\proden}   & 68.62$\pm$4.86\% & 44.47$\pm$2.33\% & 71.68$\pm$0.83\%  & 54.40$\pm$0.85\%$\bullet$ & 67.12$\pm$0.97\% \\
		{\rcpll}    & 68.89$\pm$5.02\%  & 44.59$\pm$2.65\% & 71.56$\pm$0.88\% & 54.23$\pm$0.89\%$\bullet$ & 67.04$\pm$0.88\% \\
		{\ccpll}    & 62.21$\pm$1.77\%$\bullet$  & 47.49$\pm$2.31\%& 68.42$\pm$0.99\%$\bullet$ & 53.50$\pm$0.96\%$\bullet$ & 61.92$\pm$0.96\%$\bullet$ \\
		{\dcnn}     & 68.56$\pm$6.68\%  & 43.27$\pm$2.98\%$\bullet$ & 65.48$\pm$2.57\%$\bullet$& 48.16$\pm$0.62\%$\bullet$ & 52.46$\pm$1.71\%$\bullet$ \\
		{\gaccl}    & 50.21$\pm$3.62\%$\bullet$  & 30.91$\pm$4.31\%$\bullet$ & 34.57$\pm$3.41\%$\bullet$& 50.65$\pm$0.94\%$\bullet$ & 45.72$\pm$1.75\%$\bullet$ \\
		\midrule
		{\clpl}     & \textbf{74.15$\pm$3.03\%}  & 44.47$\pm$2.58\% & 65.76$\pm$1.19\%$\bullet$& 50.01$\pm$1.03\%$\bullet$ & 53.25$\pm$1.12\%$\bullet$ \\
		{\plsvm}    & 71.56$\pm$2.71\%  & 38.25$\pm$3.89\%$\bullet$ & 50.66$\pm$4.23\%$\bullet$ & 36.39$\pm$1.03\%$\bullet$ & 51.24$\pm$0.72\%$\bullet$ \\
		{\plknn}    & 33.87$\pm$2.48\%$\bullet$ & 43.28$\pm$2.35\%$\bullet$ & 64.34$\pm$0.75\%$\bullet$ & 49.24$\pm$1.23\%$\bullet$ & 40.38$\pm$0.37\%$\bullet$ \\
		{\ipal}     & 72.10$\pm$2.75\%  & \textbf{52.96$\pm$1.36\%}$\circ$ & 70.32$\pm$0.91\%$\bullet$& 54.41$\pm$0.68\%$\bullet$ & 66.04$\pm$0.85\%$\bullet$ \\
		{\plle}     & 72.55$\pm$3.55\%  & 47.54$\pm$1.96\% & 70.63$\pm$1.24\%$\bullet$& 53.38$\pm$1.03\%$\bullet$ & 59.45$\pm$0.43\%$\bullet$  \\
		\bottomrule
	\end{tabular}
\end{table}

\subsection{Baselines}
The performance of {\vpll} is compared against five DNN based  approaches: 1) {\proden} \cite{lv2020progressive}: A progressive identification partial label learning approach which approximately  minimizes a  risk estimator and  identifies the true labels in a seamless manner;  2) {\rcpll} \cite{feng2020provably}: A risk-consistent partial label learning approach which employs the importance reweighting strategy to converges the true risk minimizer; 3) {\ccpll} \cite{feng2020provably}: A classifier-consistent partial label learning approach which uses a transition matrix to form an empirical risk estimator; 4) {\dcnn} \cite{yao2020deep}: A deep partial label learning approach which design an entropy-based regularizer to maximize the margin between the potentially correct label and the unlikely ones; 5) {\gaccl} \cite{ishida2019complementary}: An  unbiased risk estimator  approach which can be applied  for  partial label learning.

For all the DNN based approaches, we adopt the same  predictive model for fair comparisons. Specifically, the 32-layer ResNet is trained on \texttt{CIFAR-10} in which the learning rate, weight decay and mini-batch size are set to $0.05$, $10^{-3}$ and $256$, respectively. The three-layer MLP is trained on  \texttt{MNIST}, \texttt{Fashion-MNIST} and \texttt{Kuzushiji-MNIST} where  the learning rate, weight decay and mini-batch size are set to $10^{-2}$, $10^{-4}$ and $256$, respectively.  The linear model is trained  on UCI and real-world PLL datasets where the learning rate, weight decay and mini-batch size are set to $10^{-2}$, $10^{-4}$ and $100$, respectively.    We implement the comparing methods with PyTorch. The number of epochs is set to 500, in which the first 10 epochs are warm-up training. We also want to use {\vpll} on MindSpore \footnote{https://www.mindspore.cn/}, which is a new deep learning computing framework. These problems are left for future work.
 
In addition, we also compare  with  five classical partial label learning approaches, each configured with parameters suggested in respective literatures: 1) {\clpl} \cite{Cour_2011}: A convex partial label learning approach which uses averaging-based disambiguation; 2) {\plknn} \cite{Hullermeier_Beringer2006}: An instance-based partial label learning approach which works by   $k$-nearest neighbor weighted voting; 3) {\plsvm} \cite{Nguyen_Caruana2008}: A maximum margin partial label learning approach which
works by identification-based disambiguation; 
4) {\ipal} \cite{zhang2017disambiguation}: A non-parametric method that applies the label propagation strategy  to
iteratively update the confidence of each candidate label; 5) {\plle} \cite{xu2019partial}: A two-stage partial label learning approach which estimates the generalized description degree of each class label values via graph Laplacian and  induces a multi-label predictive model with the generalized description degree  in separate stages.

\subsection{Experimental Results}
Table \ref{bench_feature}   reports the classification accuracy of each DNN-based method  on benchmark datasets corrupted by the instance-dependent generating procedure.   The best results are highlighted in bold. In addition, $\bullet$ / $\circ$ indicates whether {\vpll} is statistically superior/inferior to the comparing approach on each dataset (pairwise $t$-test at 0.05 significance level).  From the  table, we can observe that {\vpll} always achieves the best performance and significantly outperforms other compared methods in most cases.  In addition, we also validate the effectiveness of our approach on  uniform corrupted versions that is commonly adopted in previous works. From Table \ref{benchmark_uniform}, we can observe that  {\vpll} achieves superior or at least comparable performance  to other approaches on  uniform corrupted versions.

Table \ref{UCI_feature}  and Table \ref{UCI_uniform} report the classification accuracy of each method on  UCI datasets corrupted by the instance-dependent generating procedure and the uniform generating procedure, respectively. {\vpll} always achieves the best performance and significantly outperforms other DNN-based  methods in most cases on instance-dependent corrupted versions while  achieves superior or at least comparable performance  to other approaches on  uniform corrupted versions. We further compare {\vpll} with five classical PLL methods that can hardly be implemented by DNNs on large-scale datasets. Despite the small scale of  most UCI datasets, {\vpll} always achieve the best performance in most cases against the classical PLL methods as {\vpll} can deal with   the high  average number of candidate labels (can be seen  in Appendix A.3) in the corrupted UCI datasets.

Table \ref{real-world datasets} reports the experimental results on real-world PLL datasets.  We can find that {\vpll}  achieves best  performance against other  DNN-based  methods  on the real-world PLL datasets. Note that {\vpll} achieves best  performance  against classical  methods on all datasets except \texttt{Lost} and \texttt{MSRCv2} as these datasets are small-scale and  the   average number of candidate labels in each dataset is low (can be seen  in Appendix A.3), which leads to the result that  DNN-based methods cannot take full advantage.

\begin{figure*}
	\centering
	\subfigure[{ Instance-dependent version}]{\label{fig:feature}\includegraphics[width=0.32\textwidth]{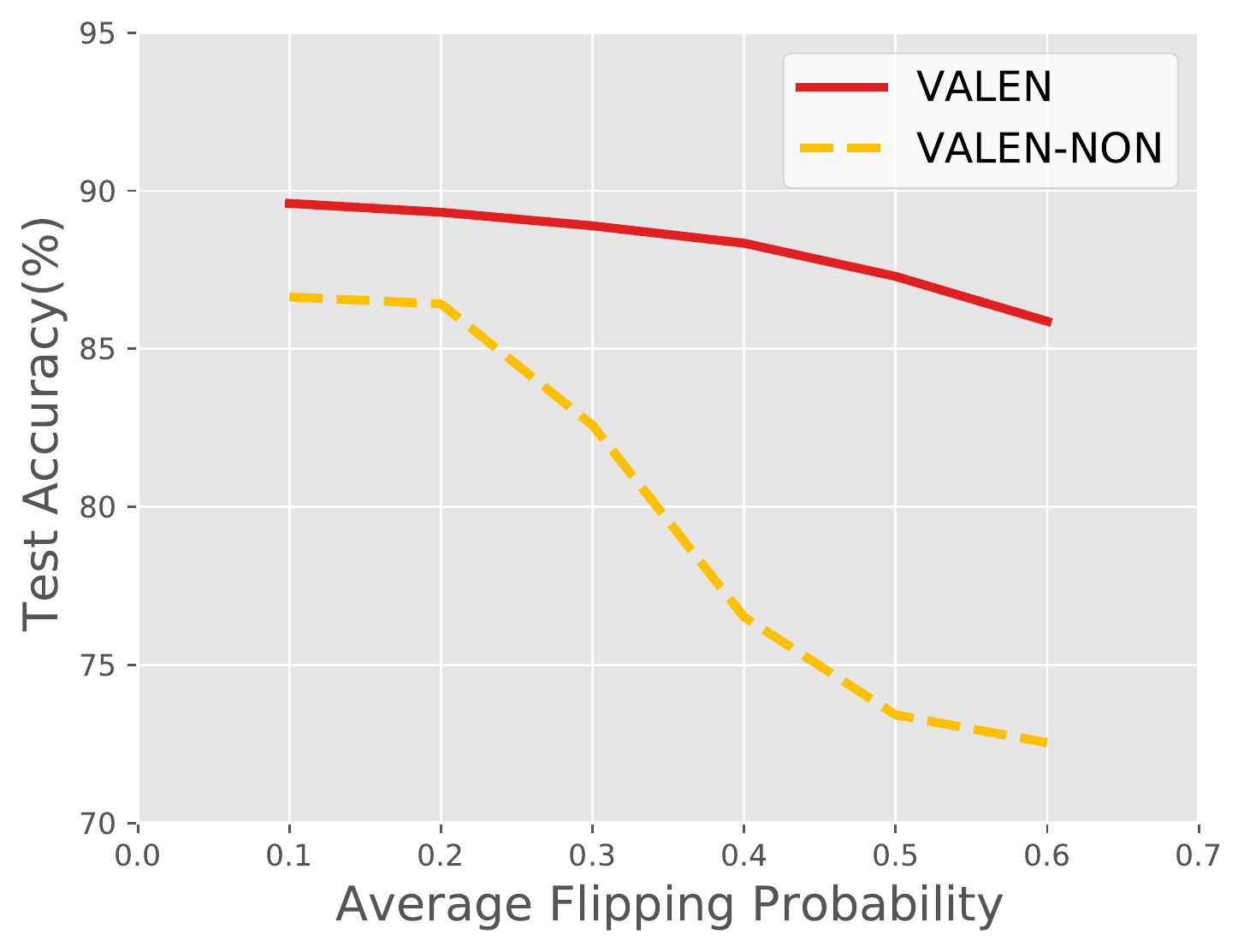}}
	\subfigure[Uniform  version]{\label{fig:uniform}\includegraphics[width=0.32\textwidth]{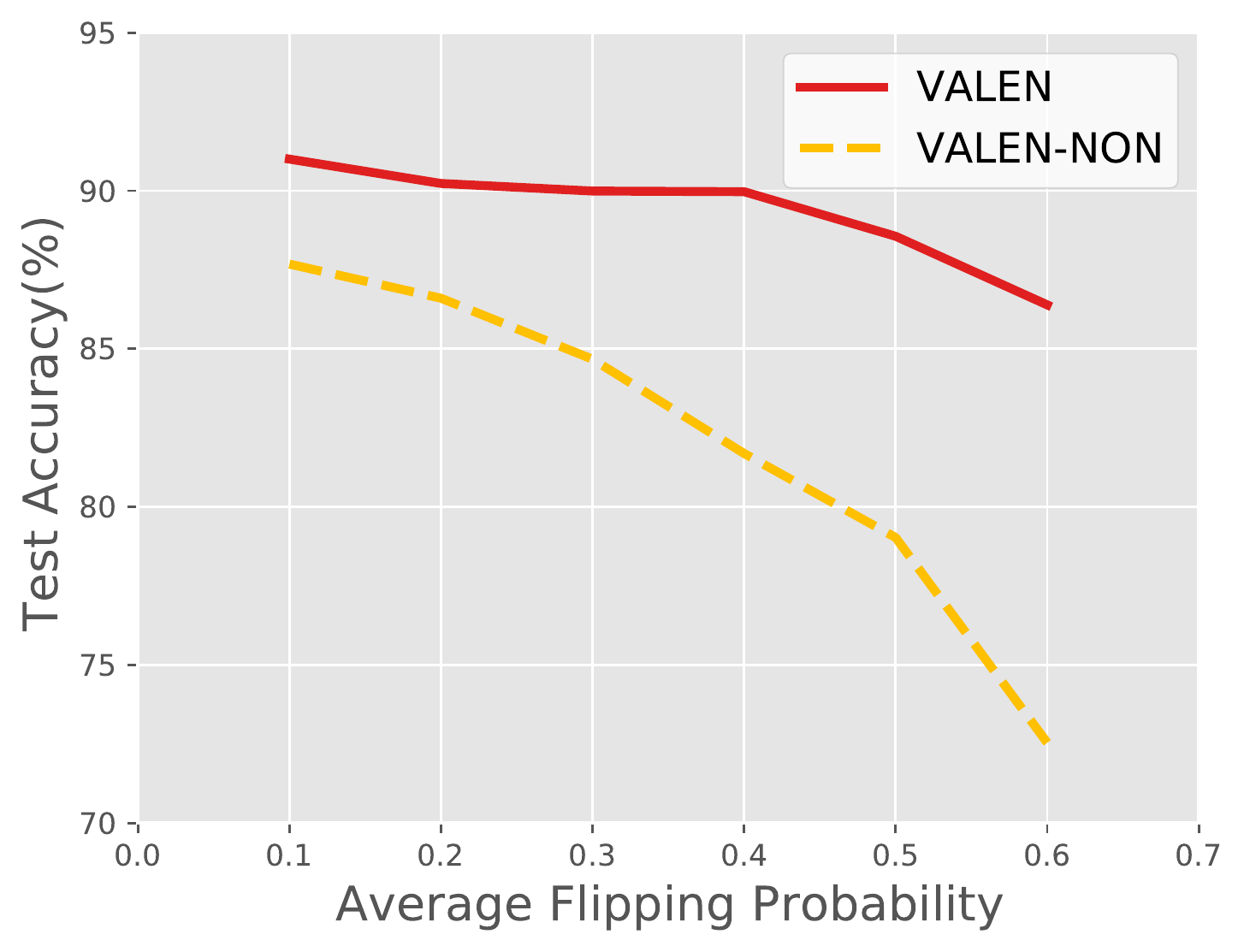}}
	\subfigure[Convergence curves of $\mathbf{D}$]{\label{fig:convergence}\includegraphics[width=0.32\textwidth]{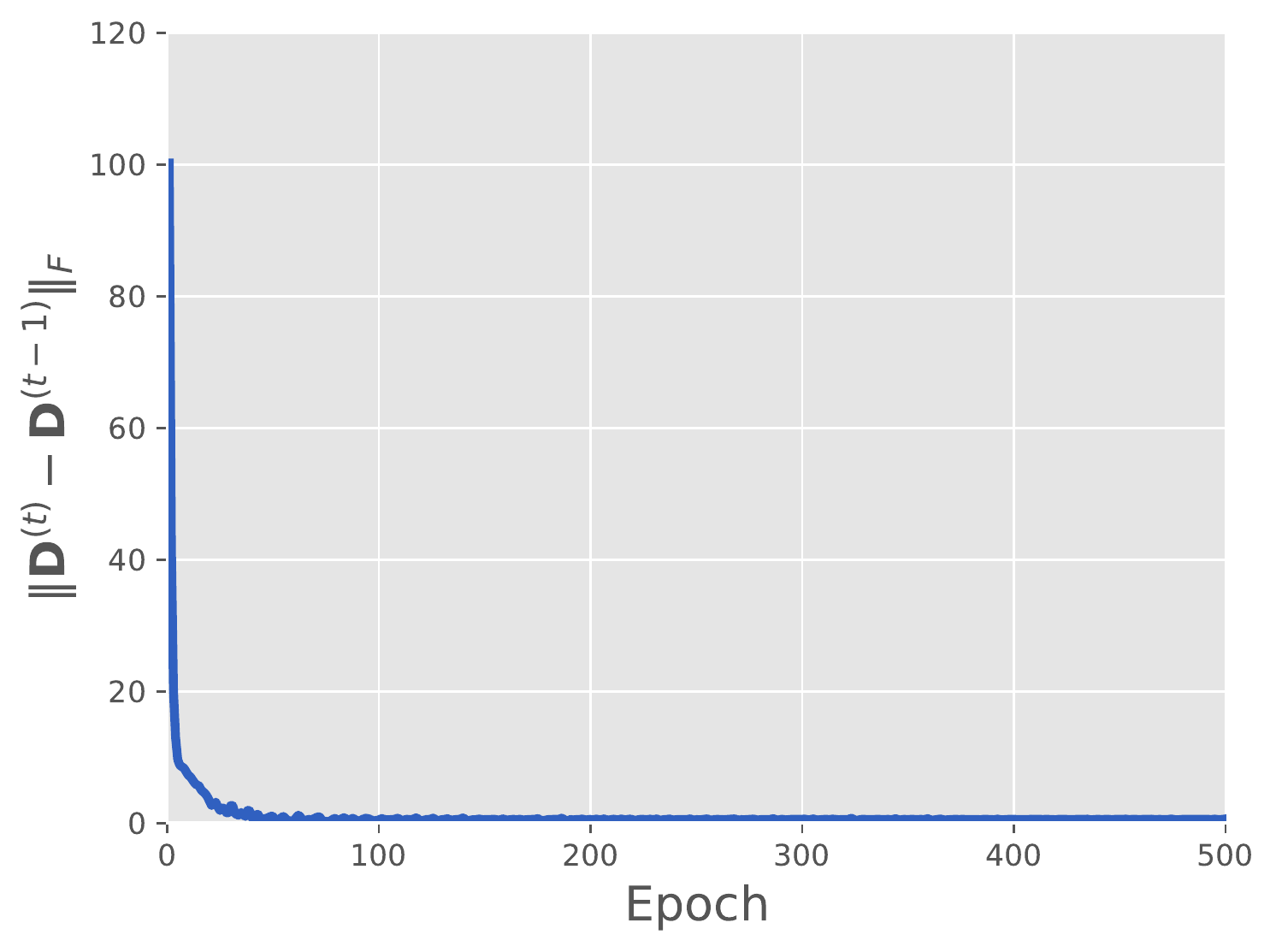}}
	\caption{Further analysis  of {\vpll} on \texttt{KMNIST}. }
\end{figure*}

Figure \ref{fig:feature} and Figure \ref{fig:uniform} illustrate the performance of {\vpll} on \texttt{KMNIST} corrupted by the instance-dependent generating procedure  and the uniform generating procedure  under different flipping probability, respectively.   Besides, the performance of the ablation version that  removes the label enhancement and trains the predictive model with PLL minimal loss (denoted by {\vpllnon}) is recorded.  These results clearly validate the usefulness of recovered label distributions for improving predictive performance. Figure \ref{fig:convergence} illustrates the recovered label distribution matrix over all training examples converges as the number of epoch (after warm-up training) on \texttt{Kuzushiji-MNIST}.  We can see that the recovered label distributions converge fast with the increasing number of epoch.

\section{Conclusion}
In this paper, the problem of partial label learning is studied  where  a  novel  approach  {\vpll} is  proposed. We  for the first time consider the instance-dependent PLL  and assume that each  partially labeled  example is associated with a latent {label distribution}, which is the essential labeling information and worth  being recovered for predictive model training. {\vpll} recovers the latent label distribution via inferring the   true posterior density of the latent label distribution by Dirichlet density parameterized with an inference model and deduce the evidence lower bound  for optimization. In addition, {\vpll} iteratively  recovers latent label distributions  and trains the predictive model  in every epoch. The effectiveness of the proposed approach is validated via comprehensive experiments on both synthesis datasets   and real-world PLL datasets.

\bibliographystyle{plainnat}
\bibliography{xning}
\end{document}